# FAST UNSUPERVISED AUTOMATIC CLASSIFICATION OF SEM IMAGES OF CD4$^+$ CELLS WITH VARYING EXTENT OF HIV VIRION INFECTION

**John M. Wandeto+* and Birgitta Dresp-Langley***

*+Department of Information Technology, Dedan Kimathi University of Technology, Kenya and *ICube Lab, UMR 7357 CNRS-Strasbourg University, France*

**Abstract**

Archiving large sets of medical or cell images in digital libraries may require ordering randomly scattered sets of image data according to specific criteria, such as the spatial extent of a specific local color or contrast content that reveals different meaningful states of a physiological structure, tissue, or cell in a certain order, indicating progression or recession of a pathology, or the progressive response of a cell structure to treatment. Here we used a Self Organized Map (SOM)-based, fully automatic and unsupervised, classification procedure described in our earlier work and applied it to sets of minimally processed grayscale and/or color processed Scanning Electron Microscopy (SEM) images of $CD4^+$ T-lymphocytes (so-called *helper cells*) with varying extent of HIV virion infection. It is shown that the quantization error in the SOM output after training permits to scale the spatial magnitude and the direction of change (+ or -) in local pixel contrast or color across images of a series with a reliability that exceeds that of any human expert. The procedure is easily implemented and fast, and represents a promising step towards low-cost automatic digital image archiving with minimal intervention of a human operator.

**Introduction**

Digital image archiving for public libraries, especially in economically disadvantaged parts of the world, will benefit from openly accessible and affordable computer algorithms that allow sorting large sets of image data automatically, without human intervention in the process, on the basis of specific contents or criteria. In the case of medical or cell images, for example, it may be required to order whole sets of image data according to the spatial extent of specific local color/contrast contents, revealing different states of a cell in a certain order, indicating progression or recession of a pathology, or of the progressive response of the cell structure to in-vivo treatment. In our previous work (Wandeto et al., 2016, 2017, 2018; Dresp-Langley et al., 2018b, 2018 c; Wandeto & Dresp-Langley, 2019), we had shown that this can be achieved by exploiting the Quantization Error (*QE*) in the output of a Self-Organized neural network Map (*SOM*) with minimal functional architecture as a highly reliable and statistically significant measure of local changes in contrast or color data in random-dot, medical, satellite, and other images. The SOM is easily implemented, learns the pixel structure of any target image in about two seconds by unsupervised "winner-take-all" feature learning, and detects local changes in contrast or color in a series of subsequent input images with a to-the-single-pixel precision, in less than three seconds for a set of 20 images. The QE in the SOM output permits to scale the spatial magnitude and the direction of change (+ or -) in a local pixel contrast/color with a reliability that exceeds that of any human expert. Here, we applied this method to the automatic classification of Scanning Electron Microscopy (SEM) images of $CD4^+$ T-lymphocytes (so-called *helper cells*) with varying extent of HIV virion infection.

**Materials and Methods**

*SEM images*

A total of 34 images were fed into SOM to generate QE output after learning. The images were minimally preprocessed to ensure equality in image size, relative pixel scale, and relative luminance contrast across images of a given series. A first image series consisted of 17 achromatic SEM images with varying extent of HIV-virion infection of the membrane of the cell represented in the different images. In these images, the virion buds are highlighted by "bundles" on the cell surface that have higher relative luminance contrast compared with that of the healthy cell membrane tissue (Figure 1). A second image series displayed color-enhanced variants of these 17 images, with healthy cell tissue coded in blue, and HIV-virions budding on the cell membrane coded in yellow (Figure 2).

*SOM training*

The SOM was trained on one of the images from each of the two series using unsupervised winner-take-all learning. The functional architecture of the SOM was minimal, represented by a fully connected neural network structure with a constant number of neurons and a constant neighborhood radius. Feature learning of the target image (SOM training), and subsequent SOM processing to generate QE output for classification of the 16 other images of a series took less than five seconds.

**Results**

Each of the randomly scattered images from the two series is associated with the QE from the SOM output for that given image, including the one on which the SOM was trained. To represent the image classification data for the achromatic series graphically, the QE output values from the analyses of the 17 grayscale SEM images were plotted in ascending order of magnitude (y-axis), and the image corresponding to each QE value was ranked as function of this magnitude (x-axis). These classification data are shown in Figure 3. The same principle was applied to the QE distribution from the analyses of the 17 color processed images. These classification data are shown in Figure 4. The QE-based ranking of the images from the two series after classification correctly reflects the extent of HIV-virion infection shown in each image, as illustrated in Figures 5 and 6 where the images are represented as a function of their rank order in the corresponding classification curve.

**Conclusions**

The QE from the SOM output achieved a 100% correct classification of the cell images in the order of the magnitude of HIV virion infection displayed in an image, which is reliably reflected by the relative magnitude of the QE. A human expert electron microscopist with more than 25 years of experience took 52 minutes to sort the color-enhanced micrographs in the correct order, and 69 minutes to roughly sort the original grayscale images, but not with a 100% correct. Moreover, for the human expert to be able to perform such a task at all, it is necessary to visualize all the images of a given series simultaneously on the computer screen, using software that allows zooming in and out of single images for close-up two-by-two comparisons.

**Figures and legends**

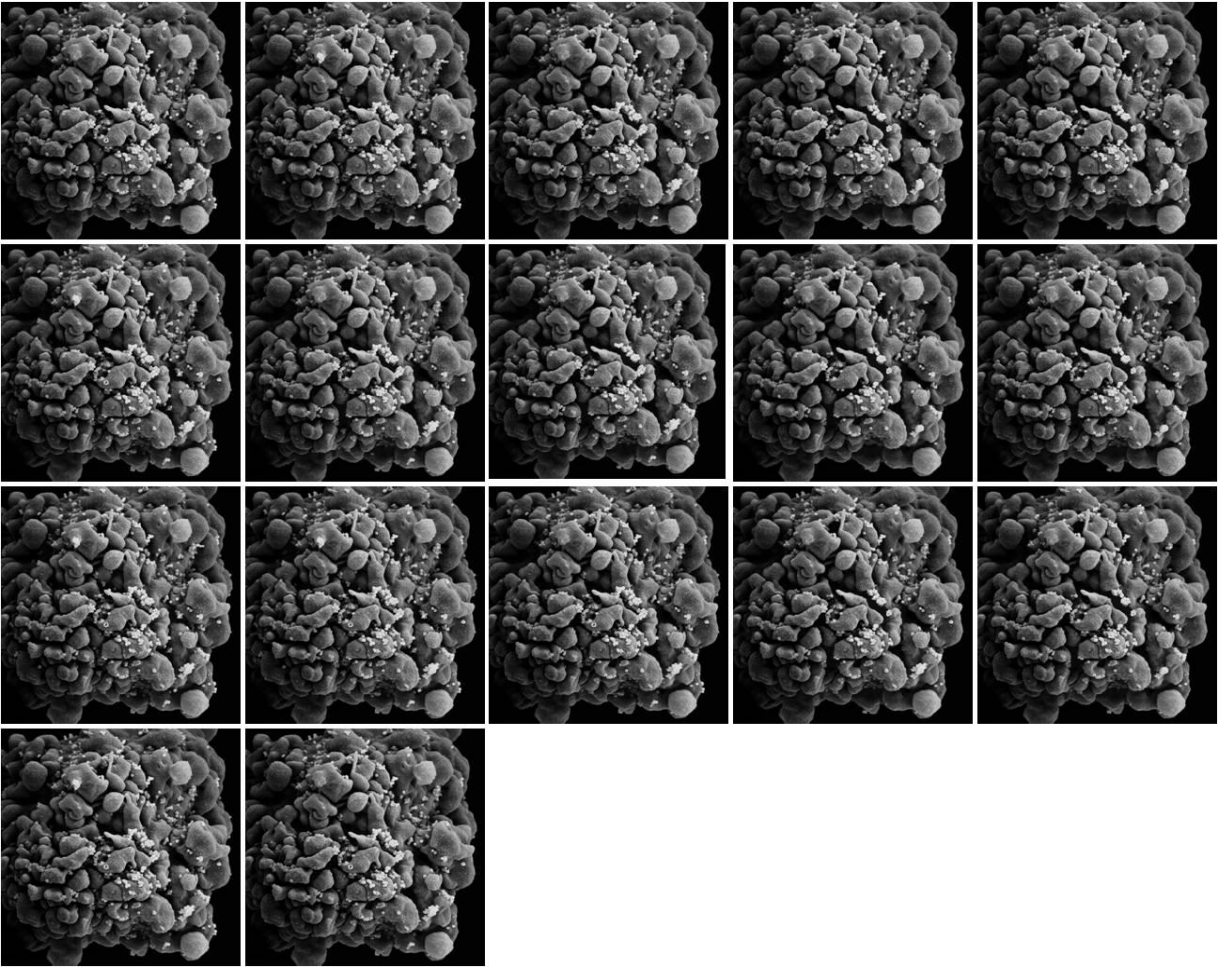

**Figure 1**

Grayscale SEM images of CD4$^+$ cells with varying extent of HIV-infection, displayed in random order before image classification

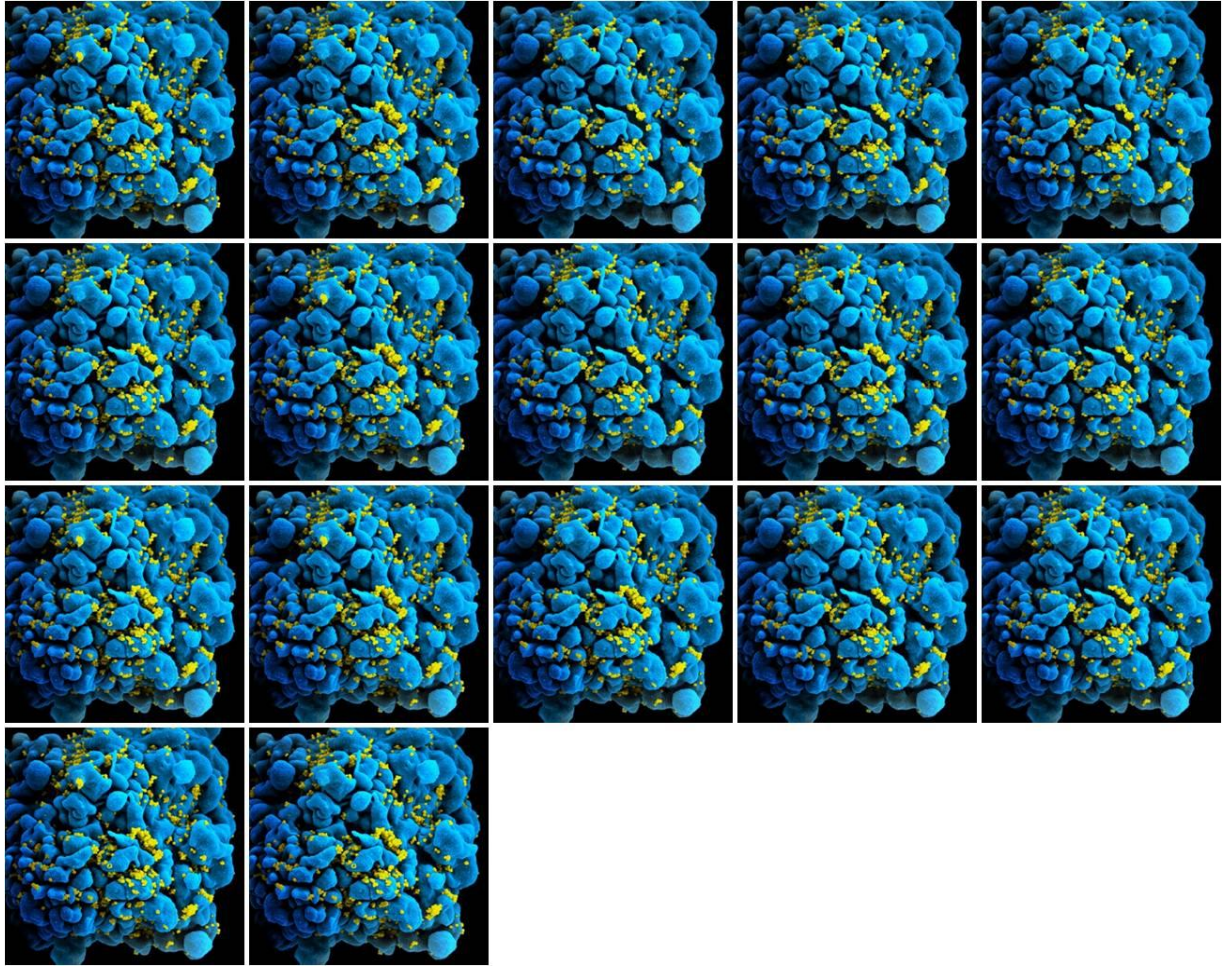

**Figure 2**

Colour coded SEM images of the same CD4$^+$ cells in random order before image classification. HIV virion bundles on the cell membrane are shown in yellow, healthy cell tissue is shown in blue

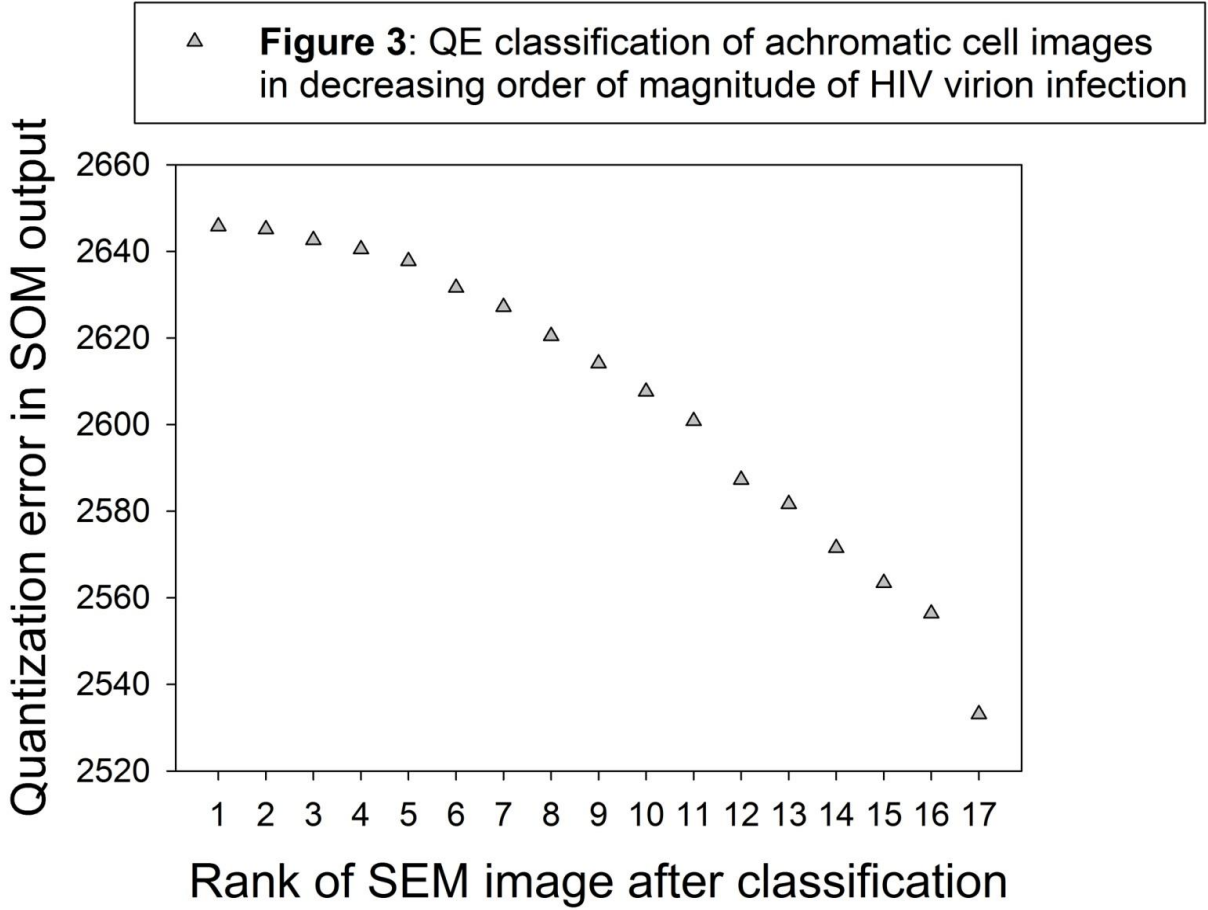

**Figure 3**

QE output values (y-axis) in decreasing order of magnitude associated with the corresponding image from the achromatic SEM image series; each image is given a classification rank number (x-axis)

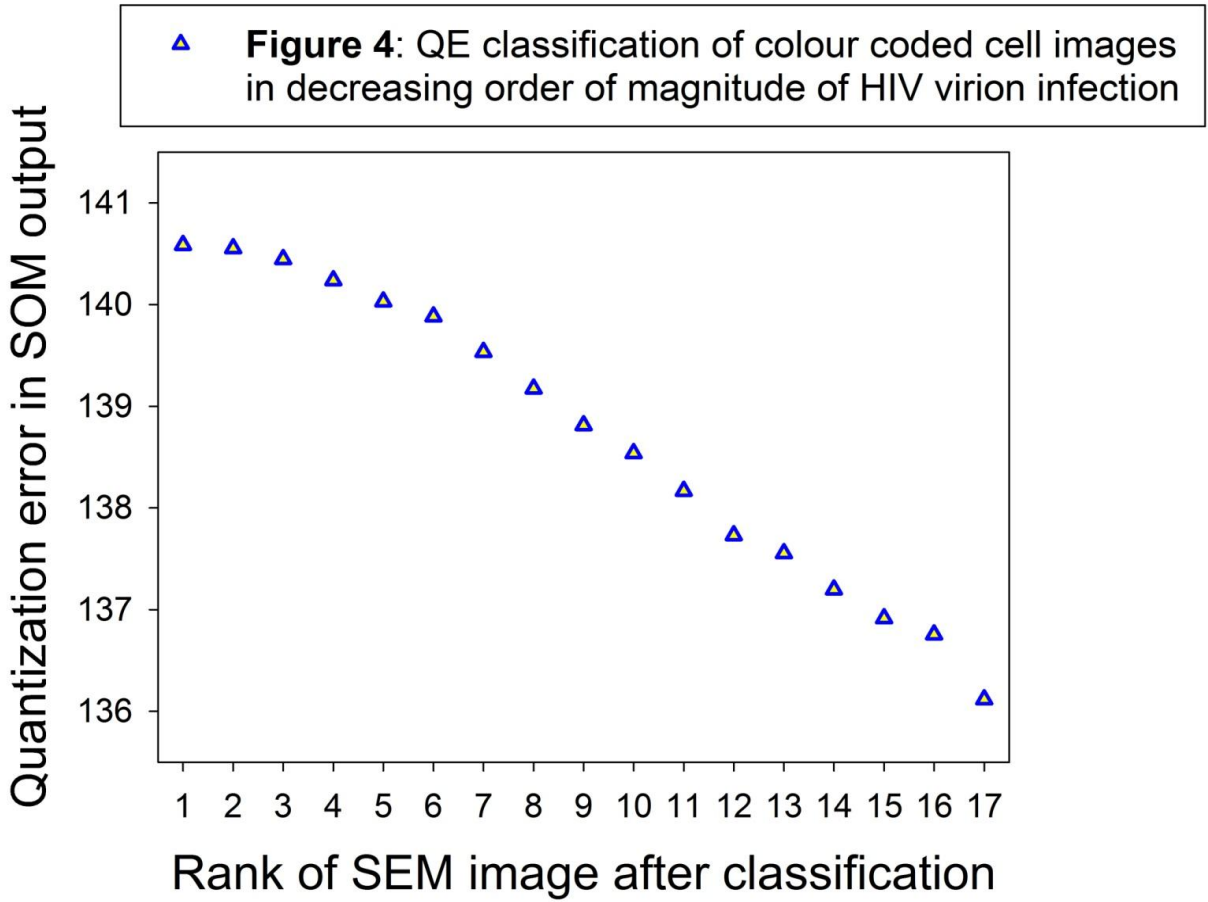

**Figure 4**

QE output values (y-axis) in decreasing order of magnitude associated with the corresponding image from the color processed SEM image series; each image is given a classification rank number (x-axis)

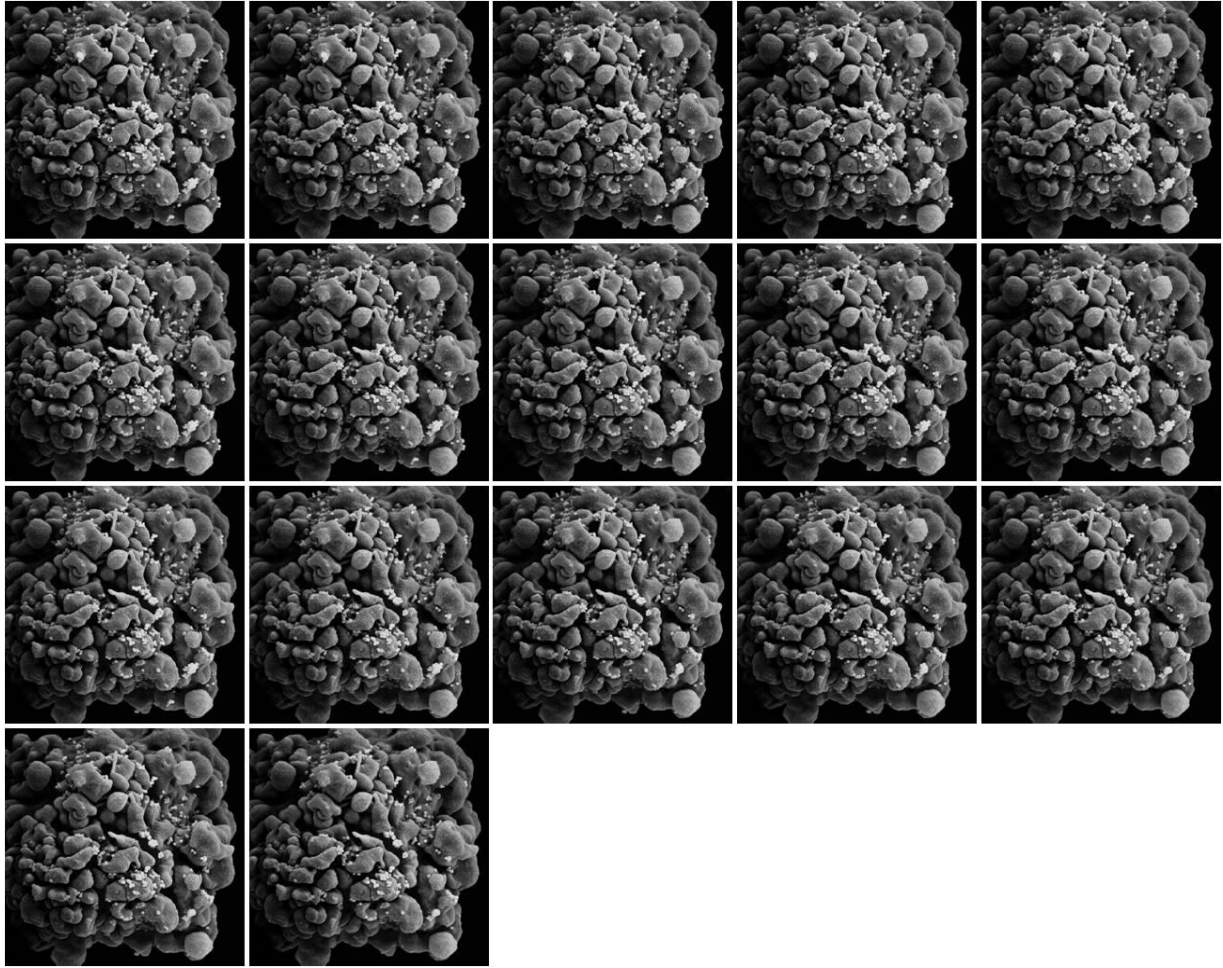

**Figure 5**

The grayscale SEM images of CD4[+] cells with gradually increasing extent of HIV-infection, displayed here in the rank order given (Figure 3) after image classification

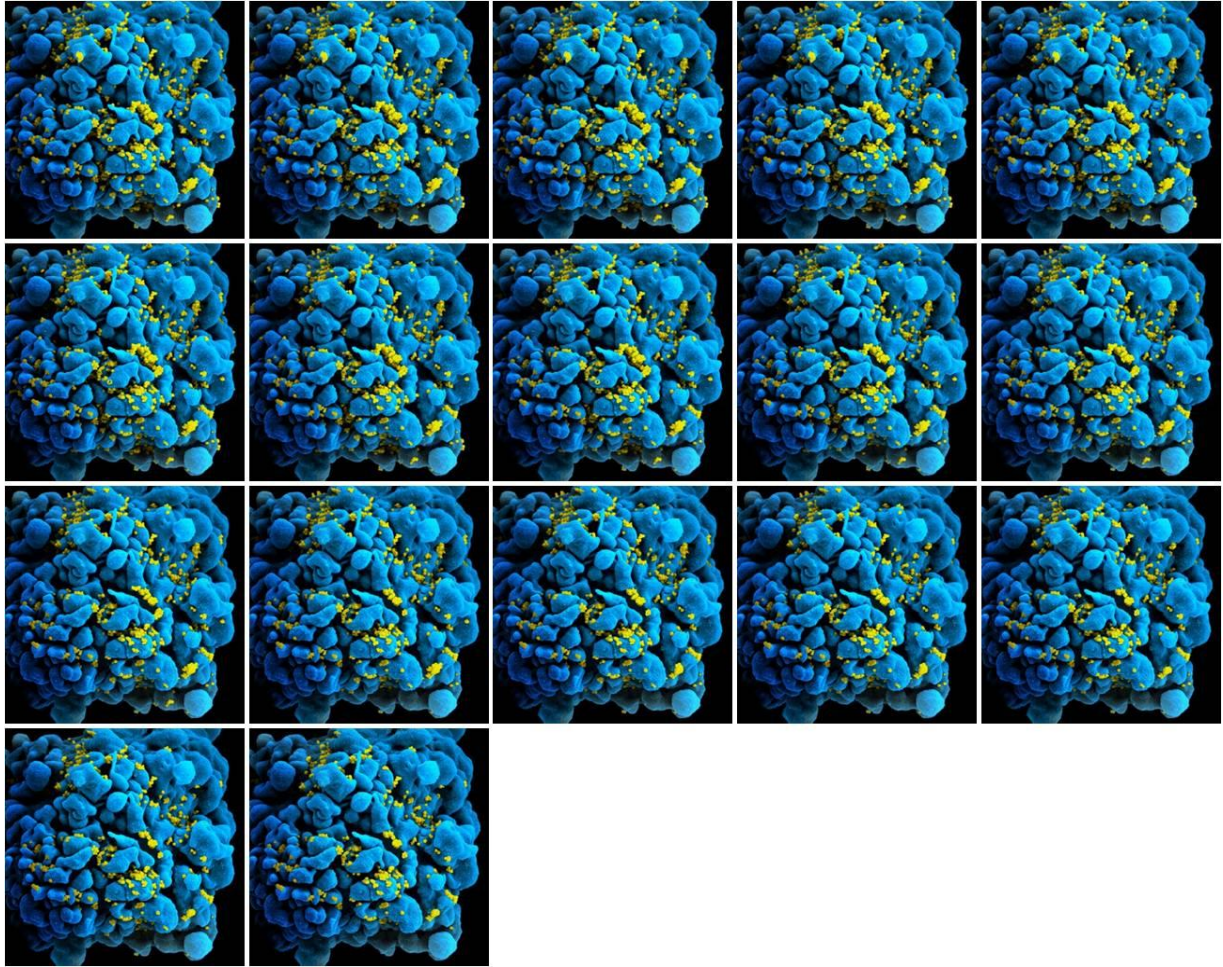

**Figure 6**

The color processed SEM images of CD4$^+$ cells with gradually increasing extent of HIV-infection, displayed here in the rank order given (Figure 4) after image classification